\documentclass[letterpaper]{article} 
\usepackage{aaai23}  
\usepackage{times}  
\usepackage{helvet}  
\usepackage{courier}  
\usepackage[hyphens]{url}  
\usepackage{graphicx} 
\usepackage{natbib}  
\usepackage{caption} 
\usepackage{algorithm}
\usepackage{algorithmic}

\usepackage{newfloat}
\usepackage{listings}
\usepackage{amssymb}
\usepackage{booktabs}
\usepackage{makecell}

\DeclareCaptionStyle{ruled}{labelfont=normalfont,labelsep=colon,strut=off} 
\floatstyle{ruled}
\newfloat{listing}{tb}{lst}{}
\floatname{listing}{Listing}
\title{ESPT: A Self-Supervised Episodic Spatial Pretext Task \\ for Improving Few-Shot Learning}
\author{
    Yi Rong\textsuperscript{\rm 1,2,3,4},
    Xiongbo Lu\textsuperscript{\rm 1},
    Zhaoyang Sun\textsuperscript{\rm 1},
	Yaxiong Chen\textsuperscript{\rm 1,2},
	Shengwu Xiong\textsuperscript{\rm 1,2,3,4}\thanks{Corresponding Author}
}
\affiliations{
    \textsuperscript{\rm 1}School of Computer Science and Artificial Intelligence, Wuhan University of Technology, Wuhan 430070, China\\
    \textsuperscript{\rm 2}Sanya Science and Education Innovation Park, Wuhan University of Technology, Sanya 572000, China\\
    \textsuperscript{\rm 3}Hainan Yazhou Bay Seed Laboratory, Sanya 572025, China\\
	\textsuperscript{\rm 4}Shanghai Artificial Intelligence Laboratory, Shanghai 200240, China \\
    \{yrong, luxiongbo, zhaoyangsun, chenyaxiong, xiongsw\}@whut.edu.cn
}

\usepackage{bibentry}

\begin{document}

\maketitle

\begin{abstract}
	Self-supervised learning (SSL) techniques have recently been integrated into the few-shot learning (FSL) framework and have shown promising results in improving the few-shot image classification performance. However, existing SSL approaches used in FSL typically seek the supervision signals from the global embedding of every single image. Therefore, during the episodic training of FSL, these methods cannot capture and fully utilize the local visual information in image samples and the data structure information of the whole episode, which are beneficial to FSL. To this end, we propose to augment the few-shot learning objective with a novel self-supervised Episodic Spatial Pretext Task (ESPT). Specifically, for each few-shot episode, we generate its corresponding transformed episode by applying a random geometric transformation to all the images in it. Based on these, our ESPT objective is defined as maximizing the local spatial relationship consistency between the original episode and the transformed one. With this definition, the ESPT-augmented FSL objective promotes learning more transferable feature representations that capture the local spatial features of different images and their inter-relational structural information in each input episode, thus enabling the model to generalize better to new categories with only a few samples. Extensive experiments indicate that our ESPT method achieves new state-of-the-art performance for few-shot image classification on three mainstay benchmark datasets. The source code will be available at: https://github.com/Whut-YiRong/ESPT.
\end{abstract}

\section{Introduction}
Deep learning \cite{DeepLearning} based approaches have achieved impressive results in various image classification tasks, such as face recognition \cite{FaceRecognition}, object recognition \cite{ObjectRecognition} and person re-identification \cite{Person}. However, the success of these methods relies heavily on the availability of massive training data with reliable annotations. Unfortunately, in many practical image classification applications, collecting and manually labeling sufficient training samples are not only expensive and time-consuming, but also may not be feasible for some rare object categories due to the scarcity of data. Training deep neural networks in such low-data regimes will inevitably lead to overfitting problems, which can greatly reduce the generalization ability of the learned models and finally limit their applicability in real-world scenarios. On the contrary, humans can rely on past experience to accurately identify new objects by only observing a small number of reference samples. By emulating such ability of human intelligence, few-shot learning has recently shown promising results in learning novel concepts from a few training images, and thus has become an effective approach to address the data scarcity problem in deep learning fields.

Few-shot learning (FSL) \cite{FSL1, FSL2, FSL3, MatchingNet, ProtoNet, MAML} aims to learn transferable prior knowledge from a set of "base classes" with sufficient training samples, and then utilize such knowledge to recognize unseen "novel classes" that only have a few reference images per class. For this purpose, the episodic learning strategy is often employed, which samples a series of few-shot episodes from the base classes in the training phase \cite{ProtoNet, MAML}. Each episode consists of a small "support set" and a relatively large "query set" that simulate the setup of the target few-shot classification tasks encountered during the evaluation procedure. Then with these episodes, a generic deep model or a common optimization method (act as the prior knowledge) is typically learned for fast adaptation to the testing few-shot classification tasks with unseen categories. One of the main problems with most FSL methods is that they usually only optimize a single categorical objective (e.g. cross-entropy loss). As a result, the learned models will only capture the necessary knowledge for the classification tasks over the training classes. Therefore, these models tend to have excessive discriminability for the base classes but limited transferability to unseen categories, which will finally lead to a decrease of few-shot classification performance on the novel classes. 

To alleviate this problem, several recent studies \cite{cc-rot, SKD, SLA, Proto-Jigsaw, InfoPatch} propose to integrate self-supervised learning (SSL) techniques into the FSL framework. They augment traditional FSL optimization objective with an annotation-free pretext task (e.g. rotation prediction or jigsaw puzzle task), which acts as an auxiliary regularization term and is jointly optimized with the classification loss. Since solving these pretext tasks depends only on the visual information present in images, the learned feature representations will be more inclined to capture low-level visual patterns in image samples, and thus will be more generalizable that can transfer well to the novel classes \cite{Contrastive}. However, existing SSL approaches typically process each input sample individually and seek the supervision signals from every single image. So the pretext task objective of these SSL methods cannot fully exploit the interrelationships of multiple image samples within a few-shot episode, which will lead to the loss of data structure information of the whole episode that may be beneficial to the FSL training process. Moreover, most SSL pretext tasks are constructed based on global image embeddings and tend to ignore the local spatial image features that contain richer and more transferable low-level visual information, which may limit their effectiveness in improving the transferability of FSL models.

In order to address the above problems, in this paper, we propose a novel self-supervised Episodic Spatial Pretext Task (ESPT) for few-shot image classification, where the supervision information is derived from the \textit{relationships between local spatial features of multiple image samples} in each learning episode. Specifically, given a few-shot episode, we first apply a random geometric transformation to all the samples in it to generate its corresponding transformed episode. These two episodes are then fed into a deep network with two identical branches, one for the original episode and the other for the transformed one. In addition to outputting label predictions for the query samples, each branch calculates the relationships between local features extracted at different spatial locations of multiple images in the input episode. After that, our ESPT objective is defined as maximizing the local spatial relationship consistency between the same images in the original and transformed episodes. The intuition behind such definition is that the transformation imposed on the images should not significantly change the local spatial relationships among them. With this objective, the introduced self-supervised pretext task is able to fully exploit the local spatial features of different images and their inter-relational structural information in the input episode. Finally, by jointly optimizing the few-shot classification loss and the proposed ESPT objective, the resulting model will benefit from the above self-supervision information to learn the visual representations with stronger transferability that can be well adapted to novel classes with few training examples, which will effectively improve its few-shot classification performance.

The main contributions of our method can be summarized as: (1) To the best of our knowledge, the proposed ESPT method is the first attempt to augment traditional few-shot image classification model using the self-supervision information obtained from the local spatial relationships among multiple image samples in each learning episode. (2) The proposed ESPT method does not introduce any additional network structures and extra trainable parameters. Therefore it does not increase the model complexity and the risk of data overfitting, which is especially important for solving few-shot learning problems with limited training samples. (3) Extensive experiments on the miniImageNet, tieredImageNet and CUB-200-2011 datasets show that our method outperforms several benchmark approaches and achieves the state-of-the-art few-shot image classification performance.

\section{Related Work}
\subsection{Few-shot Learning}
Existing few-shot classification approaches can be roughly divided into three categories: \textbf{(1) Metric/Embedding-based}  methods project input samples into a discriminative embedding space and then calculate the distance between them and the categories to be classified. MatchingNet \cite{MatchingNet} and ProtoNet \cite{ProtoNet} learn an embedding space where a predefined metric (e.g., Euclidean distance and cosine similarity) can be used as the distance measurement. However, since a common embedding space cannot be equally effective for all few-shot classification tasks, MetaOptNet \cite{MetaOptNet}, CTM \cite{CTM} and FEAT \cite{FEAT} propose to learn task-specific embeddings (or classifiers) to capture the most discriminative information for each target task. Besides, there are also some other FSL methods that design learnable distance metrics via nonlinear relation modules \cite{RelationNet, CTX, RENet}, ridge/logistic regression \cite{Baseline, Solver, RFS}, and graph neural networks \cite{GNN, CRF-GNN}. \textbf{(2) Optimization-based} methods typically meta-learn an optimizer or a model that can quickly adapt to the unseen novel classes. MAML \cite{MAML} and some of its variant methods \cite{LEO, TAML, SIB, E3BM} attempt to learn a good parameter initialization such that the adapted model of each input task can be rapidly obtained by performing only a few stochastic gradient descent (SGD) steps on support samples. In \cite{Meta-LSTM, Meta-SGD}, LSTM based meta-learners are trained to generate update rules to replace the SGD optimizer for model parameter training. \textbf{(3) Generation-based} methods aim to increase the number of training samples through data generation and augmentation. \cite{Hallucinating} and \cite{Imaginary} propose a hallucinator module that maps real training images and randomly sampled noise to hallucinated samples. The generated samples may not be realistic, but are useful to refine the decision boundary of the learned FSL model. In \cite{MetaGAN, Delta-encoder, Meta-Variance, Free-Lunch}, the intra-class variance and data distribution of the base classes are transferred to the examples in novel classes to produce augmented data.

\begin{figure*}[t]
	\centering
	\includegraphics[width=1.0 \textwidth]{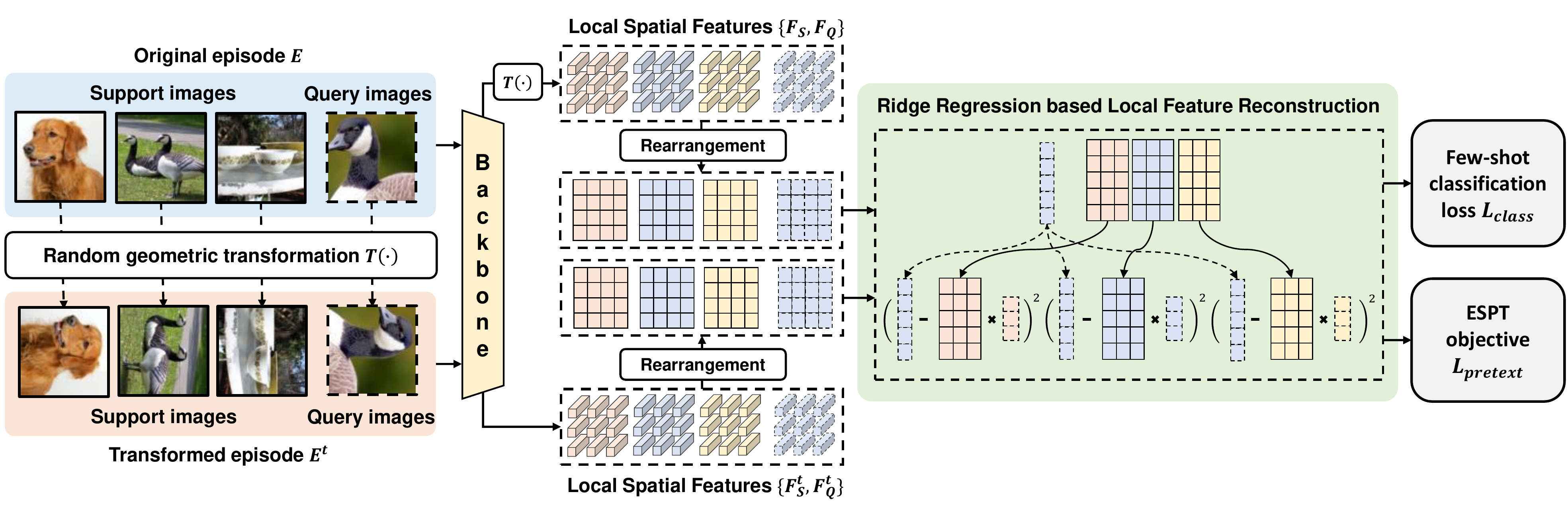} 
	\caption{Illustration of our ESPT method in the 3-way 1-shot setting. Given an input episode $ E $, we first generate a transformed episode $ E^t $ from it by using a random geometric transformation $ T(\cdot) $. Then we feed the image samples within these two episodes into a two-branch network to extract their local spatial features $ \{F_S, F_Q\} $ and $ \{F^t_S, F^t_Q\} $. After that, we establish the local spatial relationships between the support and query images from each of the two episodes by solving a ridge regression based feature reconstruction problem. Finally, for model training, the few-shot classification loss $ L_{class} $ and the proposed ESPT objective $ L_{pretext}$ are defined based on the reconstruction residuals and reconstruction coefficients, respectively.}
	\label{ESPTModel}
\end{figure*}

\subsection{SSL Augmented Few-shot Learning}
To facilitate the transferability of learned feature representations, several recent works incorporate existing SSL techniques into the FSL framework by introducing auxiliary pretext task objectives, such as predicting image rotation angles \cite{SKD, SLA}, solving image patch jigsaw puzzles \cite{Proto-Jigsaw}, estimating image patch relative locations \cite{cc-rot}, and completing contrastive learning tasks \cite{Inv-Equ, Contrastive, SCL}. However, since these pretext tasks are not specially designed for the few-shot classification problem, they may not be able to capture and utilize the episode-specific information during the episodic training process of FSL models. To this end, IEPT \cite{IEPT} designs an episode-level pretext task based on the label prediction probability of each query sample. InforPatch \cite{InfoPatch} proposes a new contrastive learning scheme that defines the positive and negative sample pairs for each anchor query image from the support samples. Different from these existing approaches, our ESPT method seeks the self-supervision signals from the local spatial relationships among multiple images in each episode. It therefore can capture and exploit the low-level visual features of image samples and the data structure information of the whole episode to learn more generalizable feature representations.

\section{Method}
\subsection{Preliminary}
In this work, we follow a standard setup of the few-shot classification problem, which is typically described as an $ n $-way $ k $-shot task. That is, in each task, there are $ n $ categories to be classified and each category contains $ k $ training samples. With this setting, few-shot learning aims to train a deep model on the data of the base classes $ C_b $, with the hope that the learned model can generalize well to the few-shot classification tasks sampled from the novel classes $ C_n $ that are not overlapped with $ C_b $, i.e., $ C_b \cap C_n = \emptyset $. For this purpose, the episodic learning strategy is employed, which constructs a series of few-shot episodes for model training. Each episode $ E = \{S, Q\} $ contains a support set $ S=\{(x_s, y_s)| y_s \in C_e, s=1,...,n \times k \} $ and a query set $ Q=\{(x_q, y_q)| y_q \in C_e, q=1,...,n \times l \} $ that are drawn from the same label space to simulate the $ n $-way $ k $-shot setting of the testing classification tasks. Here, $ S $ and $ Q $ are totally disjoint, satisfying $ S \cap Q = \emptyset $, and $ C_e $ is a set of $ n $ classes that are randomly sampled from $ C_b $. At each training iteration, the model first adapts to the input episode by performing an update using its support set $ S $. Then the performance of the resulting model is evaluated on the corresponding query set $ Q $ to produce an optimization loss that is used to update the global model parameters for all episodes.

\subsection{The Proposed Two-Branch Network}
The main purpose of our ESPT method is to augment FSL by integrating an auxiliary self-supervised pretext task, so that the learned model can benefit from class-agnostic self-supervision information to learn more transferable feature representations. To achieve this, as shown in Figure \ref{ESPTModel}, we construct a deep network with two identical branches that share the same feature extractor $ f_\theta $, which is implemented as a convolutional neural network (e.g., ResNet, WRN). For each input episode $ E=\{S, Q\} $, we first apply a random geometric transformation $ T(\cdot) $ to all the samples in it to generate its corresponding transformed episode $ E^t=\{S^t, Q^t\} $, where $ S^t=\{(T(x_s), y_s)| y_s \in C_e, s=1,...,n \times k \} $ and $ Q^t=\{(T(x_q), y_q)| y_q \in C_e, q=1,...,n \times l \} $. Then, we feed $ E $ and $ E^t $ into the two branches of our model respectively, and obtain the convolutional feature maps of their image samples through the feature extractor $ f_\theta $ as follows:
\begin{equation}
	F_S=\{(f_s, y_s)\}=\{(T(f_\theta(x_s)), y_s)\}, 
\end{equation}
\begin{equation}
	F_Q=\{(f_q, y_q)\}=\{(T(f_\theta(x_q)), y_q)\}, 
\end{equation}
\begin{equation}
	F^t_S=\{(f^t_s, y_s)\}=\{(f_\theta(T(x_s)), y_s)\}, 
\end{equation}
\begin{equation}
	F^t_Q=\{(f^t_q, y_q)\}=\{(f_\theta(T(x_q)), y_q)\}.
\end{equation}
Here we apply the same geometric transformation $ T(\cdot) $ to the feature map of each sample in the original episode $ E $ (see equations (1), (2)), ensuring that the spatial locations in the feature maps $ T(f_\theta(x)) $ and $ f_\theta(T(x)) $ of the same image are aligned. The size of each image feature map is $ h \times w \times d $, i.e., $ \{f_s, f_q, f^t_s, f^t_q \} \in \mathcal{R}^{h \times w \times d} $, where $ h $ and $ w $ denote its height and width respectively, and $ d $ is the dimension of its local feature vector at each spatial location. With these local image features, the objective function of our ESPT method is defined in the following sections.

\subsection{Episodic Spatial Pretext Task}
We construct our episodic spatial pretext task based on the local spatial relationships among multiple image samples in each training episode to effectively capture and exploit its data structure information. To establish such relationships for an input episode $ E $, we propose to reconstruct the local feature vectors of the query images by using the spatial features of the support samples. Concretely, for each class $ c \in C_e $, we first rearrange the feature maps of its $ k $ support samples ($ y_s=c $) into a single spatial feature matrix $ X_c \in \mathcal{R}^{khw \times d} $. Then, each local feature vector $(f_q)_{ij}\in \mathcal{R}^d$ of the query image $ x_q $ is reconstructed by solving the following linear least-squares problem:
\begin{equation}
	\min_{(w_q)^c_{ij}} \big\|(f_q)_{ij} - X_c^T (w_q)^c_{ij} \big\|^2_2 + \lambda \big\|{(w_q)^c_{ij}}\big\|^2_2,
\end{equation}
where $ \|\cdot\|_2 $ denotes the $ l_2 $ norm of a vector and $ (\cdot)^T $ is the transpose operator of a matrix. $ \lambda > 0 $ is a trade-off parameter that balances the importance of the regularization term. Since the size of spatial feature matrix $ X_c $ changes depending on the number of support samples $ k $ , the feature map size $ hw $ and the local feature vector dimensions $ d $, we also need to adaptively adjust the value of $ \lambda $ according these variables to guarantee the effectiveness of the reconstruction. Therefore, the parameter $ \lambda $ can be formulated as:
\begin{equation}
	\lambda = \frac{khw}{d}\bar{\lambda},
\end{equation}
we can control $ \lambda $ by setting different value of $\bar{\lambda}$. $ (f_q)_{ij} (i=1,...,h, j=1,...,w)$ is the local feature vector at the $ (i,j) $-th spatial location of $ f_q $, and $ (w_q)^c_{ij} \in \mathcal{R}^{khw} $ denotes its reconstruction coefficients corresponding to $ X_c $. Therefore, $ (w_q)^c_{ij} $ can be used to represent the local spatial relationships between $ (f_q)_{ij} $ and the support images of the $ c $-th class. The above optimization problem in equation (5) is also known as the ridge regression problem, which has a differentiable closed-form solution that can be rapidly calculated as:
\begin{equation}
	(w_q)^c_{ij} = (X_c X_c^T+\lambda I)^{-1} X_c (f_q)_{ij}.
\end{equation}

For the transformed episode $ E^t $, by applying the same operations, we can also get the reconstruction coefficients $ (w^t_q)^c_{ij} $ for the local feature vector $ (f^t_q)_{ij} $ of the same query image $ x_q $. With these coefficients, we promote the spatial relationship consistency between each query image in the origin episode and the transformed episode by minimizing the distance between $ (w_q)^c_{ij} $ and $ (w^t_q)^c_{ij} $ for all classes at different spatial locations in the input image, which can be formulated as the following consistency loss:
\begin{equation}
	L^q_{cons} = \frac{1}{h \times w } \sum^{h}_{i=1} \sum^{w}_{j=1} \sum_{c \in C_e} dis(sg[(w_q)^c_{ij}], (w^t_q)^c_{ij}),
\end{equation}
where $ dis(\cdot, \cdot) $ denotes a distance function (the cosine distance is used in our implementation). We perform a stop gradient operation $ sg[\cdot] $ on $ (w_q)^c_{ij} $ to prevent its discriminability from being affected. By averaging the above loss over all query images, the self-supervised objective function of the propose episodic spatial pretext task is defined as:
\begin{equation}
	L_{pretext}=\frac{1}{n \times l} \sum_{q=1}^{n \times l} L^q_{cons}.
\end{equation}

It can be seen that calculating and optimizing the above objective do not require any additional network structures and extra trainable parameters. This can make the proposed ESPT method more flexible and more suitable for solving few-shot classification problems, since it does not increase the model complexity and the risk of data overfitting.

\begin{algorithm}[tb]
	\caption{Training process of our ESPT method}
	\label{alg:algorithm}
	\textbf{Input}: The training set of the base classes $ C_b $, the transformation set $ U $, the hyperparameters $ \bar{\lambda} $ and $ \alpha $ \\
	\textbf{Output}: The learned feature extractor $ f_\theta $
	
	\begin{algorithmic}[1] 
		\STATE Initialize all learnable parameters $ \Phi=\{\gamma, \theta\}$
		\WHILE{Maximum number of iterations is not reached}
		\STATE Randomly sample an episode $ E $ from the training set and a geometric transformation $ T(\cdot) $ from $ U $
		\STATE Generate the transformed episode $ E^t $ by applying $ T(\cdot) $ to the image samples in $ E $ 
		\STATE Calculate the episodic spatial pretext task objective $ L_{pretext} $ using equation (9)
		\STATE Calculate the few-shot classification loss $ L_{class} $ using equation (12)
		\STATE Calculate the total loss $ L_{total} = L_{class} + \alpha L_{pretext} $
		\STATE Update the parameters $ \Phi $ base on $ \nabla_{\Phi}L_{total} $
		\ENDWHILE
		\STATE \textbf{return} The updated $ \Phi $
	\end{algorithmic}
\end{algorithm}

\subsection{Few-Shot Classification Loss}
For few-shot image classification, the main idea is that the residual of reconstructing the query images of the $ c $-th class with support samples of the same class ($ y_s=c $) should be much smaller than that using support images from other classes ($ y_s \neq c $). Therefore, we utilize the reconstruction error $ \|(f_q)_{ij} - X_c^T (w_q)^c_{ij} \|^2_2 $ (in equation (5)) over all feature map locations of each query image $ x_q $ to compute its prediction probability over all classes as follows:
\begin{equation}
	\langle f_q, c \rangle = -\frac{1}{h \times w } \sum^{h}_{i=1} \sum^{w}_{j=1} \|(f_q)_{ij} - X_c^T (w_q)^c_{ij} \|^2_2,
\end{equation}
\begin{equation}
	p(y=c|x_q) = \frac{\mathrm{exp}(\gamma\langle f_q, c \rangle)}{\sum_{c_i \in C_e}\mathrm{exp}(\gamma \langle f_q, c_i \rangle)},
\end{equation}
where $ \gamma $ is a trainable temperature parameter that controls the scale of probability logits. Based on these definitions, the few-shot classification loss for all the query samples in the input episode can be formulated as a cross-entropy loss:
\begin{equation}
	L_{class} = -\frac{1}{n \times l} \sum_{q=1}^{n \times l} [\mathrm{log}p(y=y_q|x_q)].
\end{equation}

\subsection{Total Loss and Full Algorithm}
By linearly combining the episodic spatial pretext task objective and the few-shot classification loss in equations (9) and (12), the total loss of the proposed ESPT method can be defined as follow:
\begin{equation}
	L_{total} = L_{class} + \alpha L_{pretext},
\end{equation}
where $ \alpha > 0 $ is a trade-off parameter for weighing the two objectives. The detailed training process of our ESPT method is summarized in Algorithm \ref{alg:algorithm}. After training, since the two branches of our model share the same parameters, we drop the branch for the transformed episodes and use the remaining one in the learned model for inference. During the evaluation phase, given an $ n $-way $ k $-shot image classification task, we calculate the prediction probability distribution for each query sample using equation (11), and classify it into the category with the highest probability value.

\section{Experiments}
\subsection{Datasets}
We verify the effectiveness of our ESPT method on three widely-used datasets for few-shot image classification, including miniImageNet \cite{MatchingNet}, tieredImageNet \cite{tieredImageNet} and CUB-200-2011 \cite{CUB}.

\textbf{miniImageNet} consists of 100 object classes randomly selected from ILSVRC-12 dataset, with each class containing 600 image samples. Using the class split in \cite{Meta-LSTM}, we take 64, 16 and 20 classes to construct the training set, validation set and testing set, respectively. 

\textbf{tieredImageNet} is a much larger subset derived from ILSVRC-12 dataset. It consists of over 779k images from 608 classes, where each class is drawn from one of 34 super-categories according to the ImageNet category hierarchy. We follow the settings in \cite{DeepEMD, RENet} by dividing this dataset into 20/351, 6/97 and 8/160 super-categories/classes for training, validation and testing, respectively. Performing few-shot image classification under this setting will be more challenging and typically requires stronger generalization ability of the classification model, since the training and testing classes are sampled from different super-categories.

\textbf{CUB-200-2011 (CUB)} is a fine-grained image dataset of different birds. It contains 11,788 image samples of 200 bird species, which are partitioned into 100 training classes, 50 validation classes and 50 testing classes, as in \cite{Baseline, FRN}.

All the images used for our experiments in the above three datasets are manually cropped and resized to 84 $ \times $ 84 pixels before input into the feature extractor.

\begin{table*}[t]
	\centering
	\begin{tabular}{@{}lccccc@{}}
		\toprule[1pt]
		&    & \multicolumn{2}{c}{\textbf{miniImageNet 5-way}}   & \multicolumn{2}{c}{\textbf{tieredImageNet 5-way}} \\
		\textbf{Model} & \textbf{Backbone} & \textbf{1-shot} & \textbf{5-shot} & \textbf{1-shot} & \textbf{5-shot} \\ \midrule \midrule
		\textbf{FSL methods w/o SSL} &  &  &  &  &  \\
		MatchingNet \cite{MatchingNet} &      ResNet-12 & 65.64$ \pm $0.20 & 78.72$ \pm $0.15 & 68.50$ \pm $0.92 & 80.60$ \pm $0.71 \\
		ProtoNet \cite{ProtoNet} &       	  ResNet-12 & 62.39$ \pm $0.21 & 80.53$ \pm $0.14 & 68.23$ \pm $0.23 & 84.03$ \pm $0.16 \\
		MetaOptNet \cite{MetaOptNet} &        ResNet-12 & 62.64$ \pm $0.61 & 78.63$ \pm $0.46 & 65.99$ \pm $0.72 & 81.56$ \pm $0.53 \\
		Baseline \cite{Baseline} &            ResNet-18 & 51.75$ \pm $0.80 & 74.24$ \pm $0.63  & - & - \\
		Baseline++ \cite{Baseline} &          ResNet-18 & 51.87$ \pm $0.77 & 75.68$ \pm $0.63  & - & - \\
		Neg-Cosine \cite{Neg-Cosine} & 	      ResNet-12 & 63.85$ \pm $0.81 & 81.57$ \pm $0.56 & - & - \\
		E$^3$BM \cite{E3BM}& 		 	  	  ResNet-12 & 64.09$ \pm $0.37 & 80.29$ \pm $0.25 & 71.34$ \pm $0.41 & 85.82$ \pm $0.29 \\
		FEAT \cite{FEAT} & 			 		  ResNet-12 & 66.78$ \pm $0.20 & 82.05$ \pm $0.14 & 70.80$ \pm $0.23 & 84.79$ \pm $0.16 \\
		RFS-simple \cite{RFS} & 	 	      ResNet-12 & 62.02$ \pm $0.63 & 79.64$ \pm $0.44 & 69.74$ \pm $0.72 & 84.41$ \pm $0.55 \\
		RFS-distill \cite{RFS} & 	 		  ResNet-12 & 64.82$ \pm $0.60 & 82.14$ \pm $0.43 & 71.52$ \pm $0.69 & 86.03$ \pm $0.49 \\
		
		Meta-Baseline \cite{Meta-Baseline} &  ResNet-12 & 63.17$ \pm $0.23 & 79.26$ \pm $0.17 & 68.62$ \pm $0.27 & 83.29$ \pm $0.18 \\
		DeepEMD$ ^\dag $ \cite{DeepEMD} & 	  ResNet-12 & 65.91$ \pm $0.82 & 82.41$ \pm $0.56 & 71.16$ \pm $0.87 & 86.03$ \pm $0.58 \\
		FRN$ ^\dag $ \cite{FRN} & 			  ResNet-12 & 66.45$ \pm $0.19 & 82.83$ \pm $0.13 & 72.06$ \pm $0.22 & 86.89$ \pm $0.14 \\
		RENet$ ^\dag $ \cite{RENet} & 		  ResNet-12 & 67.60$ \pm $0.44 & 82.58$ \pm $0.30 & 71.61$ \pm $0.51 & 85.28$ \pm $0.35 \\
		TPMN$ ^\dag $ \cite{TPMN} & 		  ResNet-12 & 67.64$ \pm $0.63 & 83.44$ \pm $0.43 & 72.24$ \pm $0.70 & 86.55$ \pm $0.63 \\ \midrule 
		\textbf{SSL augmented FSL methods} &  &  &  &  &  \\
		CC+rot \cite{cc-rot} &     	 		  WRN-28-10 & 62.93$ \pm $0.45 & 79.87$ \pm $0.33 & 70.53$ \pm $0.51 & 84.98$ \pm $0.36 \\
		SLA \cite{SLA} &				      ResNet-12 & 62.93$ \pm $0.63 & 79.63$ \pm $0.47 & - & - \\
		SCL \cite{SCL} & 			 		  ResNet-12 & 65.69$ \pm $0.81 & 83.10$ \pm $0.52 & 71.48$ \pm $0.89 & 86.88$ \pm $0.53 \\
		SKD \cite{SKD} & 			 		  ResNet-12 & 67.04$ \pm $0.85 & \textbf{83.54$ \pm $0.54} & 72.03$ \pm $0.91 & 86.50$ \pm $0.58 \\
		CPLAE \cite{CPLAE} & 	 	 		  ResNet-12 & 67.46$ \pm $0.44 & 83.22$ \pm $0.29 & 72.23$ \pm $0.50 & \textbf{87.35$ \pm $0.34} \\
		IEPT \cite{IEPT} & 			 		  ResNet-12 & 67.05$ \pm $0.44 & 82.90$ \pm $0.30 & \textbf{72.24$ \pm $0.50} & 86.73$ \pm $0.34 \\
		InfoPatch \cite{InfoPatch} & 	 	  ResNet-12 & \textbf{67.67$ \pm $0.45} & 82.44$ \pm $0.31 & 71.51$ \pm $0.52 & 85.44$ \pm $0.35 \\ \midrule 	
		ESPT$ ^\dag $ (Ours) &    			  ResNet-12 & \textbf{68.36$ \pm $0.19} & \textbf{84.11$ \pm $0.12} & \textbf{72.68$ \pm $0.22} & \textbf{87.49$ \pm $0.14} \\	
		\bottomrule[1pt]
	\end{tabular}
	\caption{Performance comparison on miniImageNet and tieredImageNet. The mean 5-way few-shot classification accuracies ($\%$, top-1) with the 95$\%$ confidence intervals are reported. $ \dag $ denotes the methods using local image representations.}
	\label{GeneralFewShot}
\end{table*}

\subsection{Implementation Details}
\subsubsection{Backbone}
For a fair comparison, the ResNet-12 network \cite{ResNet} with the same architecture as previous works \cite{Solver, FEAT} is used as the feature extractor $ f_\theta $ of our model. This ResNet-12 network consists of 4 residual blocks, each containing 3 convolutional layers with the kernel size of 3 $ \times $ 3. The number of filters in each convolutional layer of the four blocks is set to 64, 160, 320, and 640, respectively. Each residual block is followed by a 2 $ \times $ 2 max-pooling layer for down-sampling the feature maps. We remove the global average-pooling layer on top of the network to preserve the local spatial features. Therefore, for each input image with 84 $ \times $ 84 pixels, the feature extractor $ f_\theta $ will output a feature map with the size of 640 $ \times $ 5 $ \times $ 5, i.e., $ h=5 $, $ w=5 $ and $ d=640 $.

\subsubsection{Image Transformation}
During the training of our ESPT method, for each input episode, we randomly select a geometric transformation $ T(\cdot) $ from a predefined image transformation set $ U $. In this work, we define $ U $ as a collection of 2D rotation transformations with one or more rotation degrees, i.e., $ U \subset \{ 90^{\circ},180^{\circ},270^{\circ} \} $, since the spatial location alignment between feature maps $ T(f_\theta(x)) $ and $ f_\theta(T(x)) $ (see equations (1),(3) and (2),(4)) can be easily achieved under this definition. We will discuss the effect of $ U $ with different rotation transformations in subsequent ablation studies. It is worth noting that other image transformations (e.g., horizontal/vertical flipping, scaling and color jittering, etc.) can also be used in the proposed ESPT method, only if they do not change the alignment of feature map spatial locations between the original input images and the transformed ones.

\subsubsection{Training Details}
In order to stabilize the training process, same as \cite{FRN}, we rescale the extracted ResNet-12 local spatial image features by a factor of $ 1/\sqrt{640} $, which is algebraically equivalent to the prediction logit normalization technique used in existing approaches \cite{ProtoNet, DSN}. The same \textbf{stochastic gradient descent (SGD) optimizer with Nesterov momentum of 0.9 and weight decay of 5e-4} is utilized for model training on the three datasets, except that the initial learning rate is set to different values. For \textbf{miniImageNet} dataset, we first pre-train our models for 350 epochs with an initial learning rate of 0.1 and a mini-batch size of 128. The learning rate is decayed by multiplying 0.1 after 200 and 300 epochs. In the subsequent episodic learning phase, the models are fine-tuned for 400 epochs with each epoch containing 100 few-shot episodes. We initialize the learning rate as 0.001 and cut it by a factor of 10 at 200 and 300 epochs. For \textbf{tieredImageNet} dataset, the pre-training process runs for 90 epochs, also using the initial learning rate of 0.1 and the mini-batch size of 128. Such initial learning rate is decreased after every 30 epochs by a factor of 10. Similar to miniImageNet, the episodic learning process runs for 450 epochs with 100 few-shot episodes in each epoch. The learning rate is initialized as 0.001 and reduced by multiplying 0.1 at 50 and 250 epochs. For \textbf{CUB-200-2011} dataset, we episodic-train our models from scratch for 800 epochs, where each epoch consists of 100 few-shot episodes as well. The initial learning rate is set as 0.05 and decreased by a factor of 10 at 500 and 650 epochs. During training on the above three datasets, we evaluate the classification performance of learned models on the validation set after every 10 epochs, and select the best-performing model throughout the training process as our final result model.

\subsubsection{Evaluation Metric}
Same as \cite{IEPT, SCL}, we take the top-1 classification accuracy as the evaluation metric, and evaluate the performance of our method under standard 5-way 1-shot and 5-way 5-shot settings. For each experiment, we randomly sample 10,000 few-shot image classification tasks from the testing split of the dataset used, where each category to be classified contains 16 query samples. The mean top-1 accuracy and the 95$\%$ confidence intervals over these sampled tasks are calculated and reported.

\begin{table}[t]
	\centering
	\begin{tabular}{@{}lccc@{}}  
		\toprule[1pt]
		&    & \multicolumn{2}{c}{\textbf{CUB-200-2011 5-way}}  \\
		\textbf{Model} & \textbf{Backbone} & \textbf{1-shot} & \textbf{5-shot} \\ \midrule \midrule
		MatchingNet &    ResNet-18 & 73.49$ \pm $0.89 & 84.45$ \pm $0.58  \\
		ProtoNet &       ResNet-18 & 72.99$ \pm $0.88 & 86.64$ \pm $0.51 \\
		MAML &     		 ResNet-18 & 68.42$ \pm $1.07 & 83.47$ \pm $0.62 \\
		Baseline& 		 ResNet-18 & 65.51$ \pm $0.87 & 82.85$ \pm $0.55 \\
		Baseline++& 	 ResNet-18 & 67.02$ \pm $0.90 & 83.58$ \pm $0.54 \\
		RelationNet &    ResNet-18 & 68.58$ \pm $0.94 & 84.05$ \pm $0.56 \\
		Neg-Cosine &     ResNet-18 & 72.66$ \pm $0.85 & 89.40$ \pm $0.43 \\ \midrule
		FEAT &			 Conv4-64  & 68.87$ \pm $0.22 & 82.90$ \pm $0.15 \\ 
		CPLAE &          Conv4-64  & 69.77$ \pm $0.50 & 84.57$ \pm $0.33 \\
		IEPT &           Conv4-64  & 69.97$ \pm $0.49 & 84.33$ \pm $0.33 \\ \midrule
		
		ProtoNet &       ResNet-12 & 78.60$ \pm $0.22 & 89.73$ \pm $0.12 \\
		RFS-simple  &            ResNet-12 & 72.78$ \pm $0.86 & 87.24$ \pm $0.50 \\
		FEAT &			         ResNet-12 & 73.27$ \pm $0.22 & 85.77$ \pm $0.14 \\
		DeepEMD$ ^\dag $ &     	 ResNet-12 & 75.65$ \pm $0.83 & 88.69$ \pm $0.50 \\ 
		RENet$ ^\dag $ &     	 ResNet-12 & 79.49$ \pm $0.44 & 91.11$ \pm $0.24 \\
		FRN$ ^\dag $ &    		 ResNet-12 & \textbf{83.55$ \pm $0.19} & \textbf{92.92$ \pm $0.10} \\ \midrule
		ESPT$ ^\dag $ (Ours) &   ResNet-12 & \textbf{85.45$ \pm $0.18} & \textbf{94.02$ \pm $0.09} \\
		\bottomrule[1pt]
	\end{tabular}
	\caption{Performance comparison on CUB-200-2011. $ \dag $ denotes the methods using local image representations.}
	\label{FineGainedFewShot}
\end{table}

\begin{table}[t]
	\centering
	\begin{tabular}{@{}lccc@{}}
		\toprule[1pt]
		&    & \multicolumn{2}{c}{\textbf{miniImageNet$ \rightarrow $CUB}} \\
		\textbf{Model} & \textbf{Backbone} & \textbf{1-shot} & \textbf{5-shot}\\ \midrule \midrule
		MatchingNet &      ResNet-10 & 35.89$ \pm $0.51 & 51.37$ \pm $0.77 \\
		RelationNet &      ResNet-10 & 42.44$ \pm $0.77 & 57.77$ \pm $0.69 \\ \midrule
		
		ProtoNet &         ResNet-18 & - & 62.02$ \pm $0.70 \\
		MAML &      	   ResNet-18 & - & 51.34$ \pm $0.72 \\
		Baseline& 		   ResNet-18 & - & 65.57$ \pm $0.70 \\
		Baseline++& 	   ResNet-18 & - & 62.04$ \pm $0.76 \\ 
		Neg-Softmax & 	   ResNet-18 & - & 69.30$ \pm $0.73 \\\midrule
		
		ProtoNet&          ResNet-12 & 47.51$ \pm $0.72 & 67.96$ \pm $0.70 \\
		MetaOptNet &       ResNet-12 & 44.79$ \pm $0.75 & 64.98$ \pm $0.68 \\
		FEAT & 			 		  ResNet-12 & 50.67$ \pm $0.78 & 71.08$ \pm $0.73 \\
		SCL & 			 		  ResNet-12 & 49.58$ \pm $0.70 & 67.64$ \pm $0.70 \\
		SCL-Distill & 			  ResNet-12 & 50.09$ \pm $0.70 & 68.81$ \pm $0.60 \\
		FRN$ ^\dag $  & 	      ResNet-12 & 51.60$ \pm $0.21 & \textbf{72.97$ \pm $0.18} \\
		TPMN$ ^\dag $ & 		  ResNet-12 & \textbf{52.83$ \pm $0.65} & 72.69$ \pm $0.52 \\  \midrule 
		ESPT$ ^\dag $ (Ours) &    ResNet-12 & \textbf{54.14$ \pm $0.21} & \textbf{74.91$ \pm $0.18}\\	
		\bottomrule[1pt]
	\end{tabular}
	\caption{Performance comparison on the cross-domain miniImageNet$ \rightarrow $CUB setting. $ \dag $ denotes the methods using local image representations.}
	\label{CrossDomainFewShot}
\end{table}

\subsubsection{Experimental Environment}
The Pytorch framework is used for our programming implementation and all the experiments are conducted in the following environment: Intel(R) Xeon(R) Gold 5117 @2.00GHz CPU, NVIDIA A100 Tensor Core GPU, and Ubuntu 18.04.6 LTS operation system. Under the above environment settings, our method takes about 162 ms for training on each 5-way 5-shot episode and such training iteration costs about 84 ms for FRN \cite{FRN}, 97 ms for RENet \cite{RENet} and over 240000 ms for DeepEMD \cite{DeepEMD}. But for each test iteration, since Eq. (5) is only applied on original image features, the computational cost of our method is reduced to be similar to that of FRN  and RENet, all around 30 ms. Our model typically converges after 75000 iterations on CUB and 35000 iterations on miniImageNet and tieredImageNet. 

\subsection{Main Results}
To evaluate the effectiveness of the proposed ESPT approach, we compare its performance with several representative and state-of-the-art FSL methods \cite{ProtoNet, DeepEMD, FRN, RENet} as well as some recently proposed SSL augmented FSL models \cite{cc-rot, SLA, IEPT, InfoPatch}. The experiments are conducted on three different few-shot learning tasks: (1) the general image classification task on miniImageNet and tieredImageNet datasets, (2) the fine-grained image classification task on CUB-200-2011 dataset, and (3) the cross-domain miniImageNet$ \rightarrow $CUB few-shot image classification task.

\subsubsection{General Few-Shot Image Classification}
To speed up the training process, we pre-train the feature extractor $ f_\theta $ of our method before the episodic learning, as in \cite{FRN}. From the experimental results reported in Table \ref{GeneralFewShot}, we can obtain the following observations: (1) Among all FSL methods w/o SSL, the approaches using local image representations outperform the others, demonstrating the importance of capturing the local spatial information for FSL. (2) The SSL augmented FSL methods generally achieve better classification performance than the FSL methods w/o SSL, suggesting that SSL can promote FSL to learn more transferable feature representations. (3) The proposed ESPT method achieves the highest 1-shot and 5-shot classification accuracies of 68.36$ \% $, 84.11$ \% $ on miniImageNet dataset and 72.66$ \% $, 87.49$ \% $ on tieredImageNet dataset. Compared with the competing methods, ESPT obtains performance gains of at least 0.69$ \% $, 0.44$ \% $ in 1-shot setting and 0.57$ \% $, 0.14$ \% $ in 5-shot setting for the two datasets, respectively. Note that ESPT achieves such improvements without using any extra network structures (e.g., the attention module used in FEAT, RENet, SCL, IEPT or the rotation classifier used in CC+rot, SLA, SKD) or technical tricks (e.g., training with larger-way episodes or higher resolution input images). Therefore, these experimental results can show the effectiveness of our ESPT method, and also demonstrate the superiority of constructing self-supervised pretext task based on the local spatial relationships among multiple image samples in each episode.

\begin{figure*}[t]
	\centering
	\includegraphics[width=0.95\textwidth]{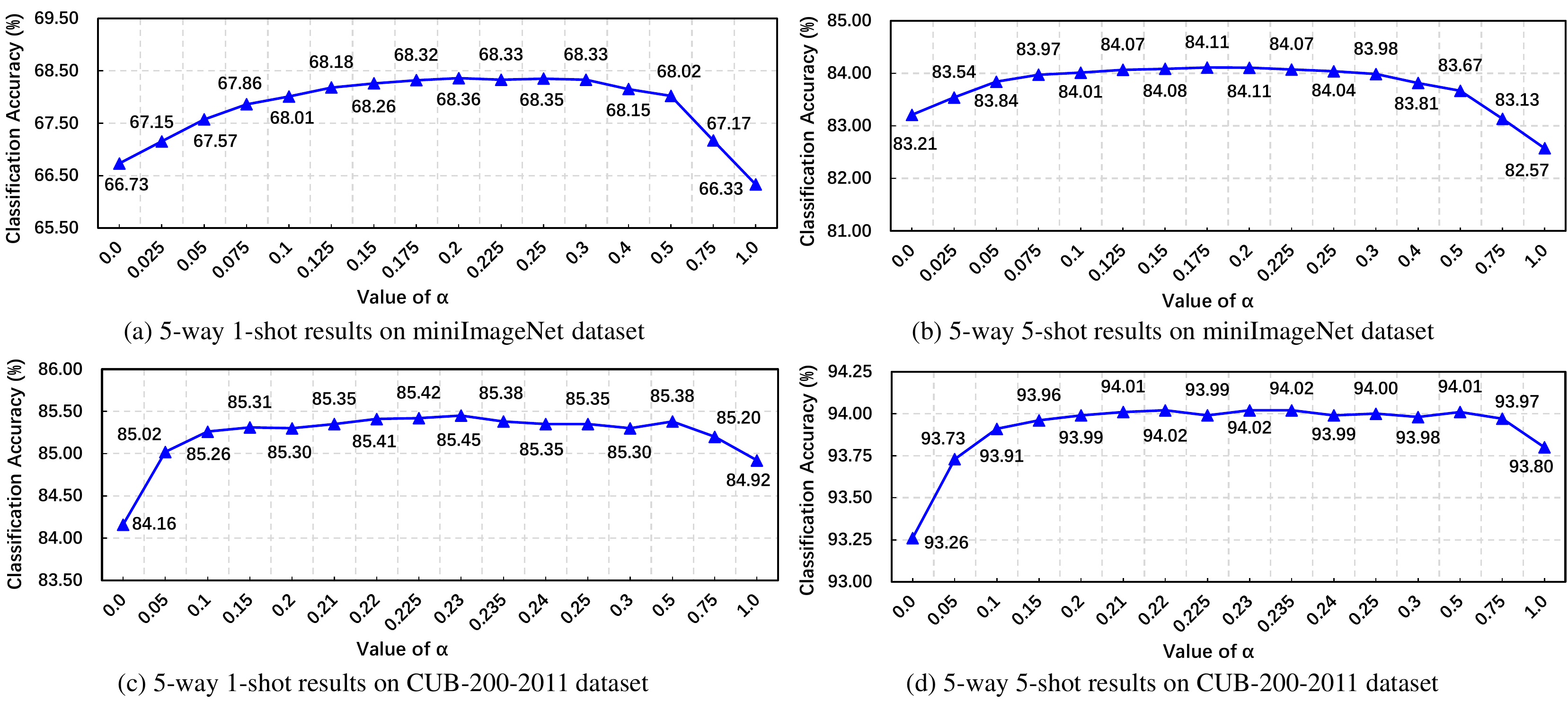}
	\caption{Effect of the proposed ESPT objective on the performance of our method on miniImageNet and CUB-200-2011.}
	\label{EffectOfESPT}
\end{figure*}

\subsubsection{Fine-Grained Few-Shot Image Classification}
 For a fair comparison, we follow the prior works \cite{Baseline, RENet, FRN} and directly episodic-train our model from scratch without using pre-training techniques. The fine-grained classification results on CUB-200-2011 dataset are presented in Table \ref{FineGainedFewShot}. It can be observed that (1) The local representation based-approaches once again achieve higher accuracies than other benchmark methods, which indicates that fine-grained FSL can also benefit from the local spatial image features. (2) Our proposed ESPT method obtains the best classification results of 85.45$ \% $ and 94.02$ \% $ for the 1-shot and 5-shot settings, respectively, outperforming the second best FRN method by a significant margin of 1.90 $ \% $ and 1.10 $ \% $, and is far superior to other competing methods. This shows that our ESPT can also be effective on the fine-grained few-shot image classification tasks. 

\subsubsection{Cross-Domain Few-Shot Image Classification} 
Following the setup in \cite{Baseline, TPMN}, we evaluate the proposed ESPT method in a more challenging cross-domain miniImageNet$ \rightarrow $CUB setting. Concretely, the model is trained on all 100 classes in the miniImageNet dataset, but validated and evaluated on 50 validation classes and 50 testing classes from the CUB-200-2011 dataset, respectively. The obtained cross-domain few-shot image classification results are shown in Table \ref{CrossDomainFewShot}. We can see that our ESPT method outperforms all benchmark approaches by a large margin. Specifically, the improvements achieved by ESPT over other competing methods range from 1.31$\%$ (vs. TPMN) to 18.25$\%$ (vs. MatchingNet) in 5-way 1-shot setting and from 1.94$\%$ (vs. FRN) to 23.57$\%$ (vs. MAML) in 5-way 5-shot setting. These experimental results indicate that augmenting FSL with our proposed ESPT objective promotes learning more transferable feature representations, which can generalize well to novel unseen categories even under domain shift.

\begin{table}[tb]
	\centering
	\begin{tabular}{p{0.5cm}<{\centering} p{0.5cm}<{\centering} p{0.5cm}<{\centering} cccc}  
		\toprule[1pt]
		\multicolumn{3}{c}{\textbf{Transformation}} & \multicolumn{2}{c}{\textbf{miniImageNet}} & \multicolumn{2}{c}{\textbf{CUB-200-2011}} \\ 
		\textbf{90$ ^{\circ} $} & \textbf{180$ ^{\circ} $} & \textbf{270$ ^{\circ} $}  & \textbf{1-shot} & \textbf{5-shot} & \textbf{1-shot} & \textbf{5-shot} \\ 
		\midrule \midrule
		$\checkmark$ &  			&  			   & 68.02 & 83.86 & 84.92 & 93.76 \\
		& $\checkmark$ &  			   & 68.19 & 84.03 & 85.06 & 93.88 \\
		& 			    & $\checkmark$ & 68.05 & 83.83 & 84.88 & 93.78 \\ \midrule
		$\checkmark$ & $\checkmark$	&  			   & 68.32 & 84.10 & 85.22 & 93.97 \\
		$\checkmark$ & 			    & $\checkmark$ & 68.18 & 84.03 & \textbf{85.45} & \textbf{94.02} \\
		& $\checkmark$	& $\checkmark$ & \textbf{68.36} & \textbf{84.11} & 85.24 & 93.95 \\	\midrule	   
		$\checkmark$ & $\checkmark$	& $\checkmark$ & \textbf{68.36} & 84.07 & 85.41 & \textbf{94.02} \\
		\bottomrule[1pt]
	\end{tabular}
	\caption{Effect of different choices of the transformation set $ U $ on the performance of our method.}
	\label{EffectOfTransformationSet}
\end{table}

\subsection{Ablation Studies}
\subsubsection{Effect of the Proposed ESPT Objective}
We analyze the effect of the proposed ESPT objective by comparing the few-shot image classification performance of our method with different loss weight (i.e., $ \alpha $ in equation (13)) values on miniImageNet and CUB-200-2011 datasets, as shown in Figure \ref{EffectOfESPT}. It can be seen that, with the value of $ \alpha $ increases from zero, the accuracy curves of our method on both datasets rise gradually at first, then reach their peak values and remain relatively stable at a high level when $ \alpha $ is within the range of $ [0.1, 0.5] $. This shows that the proposed ESPT objective can effectively and steadily improve the few-shot classification performance of our method. In addition, it can also be found that the best models of our full approach significantly outperform our method without the ESPT objective (i.e., $ \alpha=0 $). The improvements on the 5-way 1-shot and 5-way 5-shot tasks are 1.63$ \% $, 0.90$ \% $ on miniImageNet dataset and 1.29$ \% $, 0.76$ \% $ on CUB-200-2011 dataset, respectively. These results can further demonstrate the effectiveness of the proposed ESPT objective.

\subsubsection{Effect of the Transformation Set}
The 5-way classification results obtained with different transformation sets $ U $ on miniImageNet and CUB-200-2011 datasets are summarized in Table \ref{EffectOfTransformationSet}. As shown, it can be seen that the transformation sets with multiple rotation transformations typically produce better classification results than those containing only a single rotation transformation. It is mainly because that more transformations will bring more different data variants, which will benefit our ESPT method to learn feature representations with stronger generalization ability. Moreover, we can also observe that the models trained with the transformation sets $ U=\{180^{\circ}, 270^{\circ}\}$ and $ \{90^{\circ}, 270^{\circ}\}$ have the highest classification accuracies on the two datasets, respectively. We analyze the reason why $ U=\{ 90^{\circ}, 180^{\circ}, 270^{\circ} \} $ cannot lead to the best results is because that these three rotation transformations may introduce some redundant variant information that is not useful for FSL. 

\section{Conclusion}
In this paper, we augment the few-shot classification objective with a newly proposed Episodic Spatial Pretext Task (ESPT) to learn more transferable image representations. By leveraging the local spatial relationships between the support and query samples in each learning episode, ESPT can effectively capture the low-level visual information in different images and the data structure information of the whole episode, which will bring significant benefits to FSL. Extensive experiments on three widely used benchmark datasets demonstrate the effectiveness and the superiority of our ESPT method. And importantly, while achieving the state-of-the-art classification performance, the proposed ESPT method does not increase the model complexity and the risk of data overfitting, which makes ESPT more suitable for solving FSL problems with limited training data.

\section{Acknowledgments}
The research was supported by the Hainan Provincial Joint Project of Sanya Yazhou Bay Science and Technology City (Grant No: 2021JJLH0099), Project of Sanya Yazhou Bay Science and Technology City (Grant No: SCKJ-JYRC-2022-76), Postdoctoral project of Hainan Yazhou Bay Seed Laboratory (Grant No: B22E18102).

\bibliography{aaai23}

\begin{thebibliography}{52}
\providecommand{\natexlab}[1]{#1}

\bibitem[{Bertinetto et~al.(2019)Bertinetto, Henriques, Torr, and
  Vedaldi}]{Solver}
Bertinetto, L.; Henriques, J.~F.; Torr, P.; and Vedaldi, A. 2019.
\newblock Meta-learning with differentiable closed-form solvers.
\newblock In \emph{International Conference on Learning Representations}.

\bibitem[{Chen et~al.(2019)Chen, Liu, Kira, Wang, and Huang}]{Baseline}
Chen, W.-Y.; Liu, Y.-C.; Kira, Z.; Wang, Y.-C.~F.; and Huang, J.-B. 2019.
\newblock A Closer Look at Few-shot Classification.
\newblock In \emph{International Conference on Learning Representations}.

\bibitem[{Chen et~al.(2021)Chen, Liu, Xu, Darrell, and Wang}]{Meta-Baseline}
Chen, Y.; Liu, Z.; Xu, H.; Darrell, T.; and Wang, X. 2021.
\newblock Meta-Baseline: Exploring Simple Meta-Learning for Few-Shot Learning.
\newblock In \emph{Proceedings of the IEEE/CVF International Conference on
  Computer Vision (ICCV)}, 9062--9071.

\bibitem[{Doersch, Gupta, and Zisserman(2020)}]{CTX}
Doersch, C.; Gupta, A.; and Zisserman, A. 2020.
\newblock CrossTransformers: spatially-aware few-shot transfer.
\newblock In \emph{Advances in Neural Information Processing Systems},
  volume~33, 21981--21993. Curran Associates, Inc.

\bibitem[{Finn, Abbeel, and Levine(2017)}]{MAML}
Finn, C.; Abbeel, P.; and Levine, S. 2017.
\newblock Model-Agnostic Meta-Learning for Fast Adaptation of Deep Networks.
\newblock In \emph{Proceedings of the 34th International Conference on Machine
  Learning}, volume~70 of \emph{Proceedings of Machine Learning Research},
  1126--1135. PMLR.

\bibitem[{Gao et~al.(2021)Gao, Fei, Liu, Lu, and Xiang}]{CPLAE}
Gao, Y.; Fei, N.; Liu, G.; Lu, Z.; and Xiang, T. 2021.
\newblock Contrastive prototype learning with augmented embeddings for few-shot
  learning.
\newblock In \emph{Proceedings of the Thirty-Seventh Conference on Uncertainty
  in Artificial Intelligence}, volume 161 of \emph{Proceedings of Machine
  Learning Research}, 140--150. PMLR.

\bibitem[{Gidaris et~al.(2019)Gidaris, Bursuc, Komodakis, Perez, and
  Cord}]{cc-rot}
Gidaris, S.; Bursuc, A.; Komodakis, N.; Perez, P.; and Cord, M. 2019.
\newblock Boosting Few-Shot Visual Learning With Self-Supervision.
\newblock In \emph{Proceedings of the IEEE/CVF International Conference on
  Computer Vision (ICCV)}, 8059--8068.

\bibitem[{Hariharan and Girshick(2017)}]{Hallucinating}
Hariharan, B.; and Girshick, R. 2017.
\newblock Low-Shot Visual Recognition by Shrinking and Hallucinating Features.
\newblock In \emph{Proceedings of the IEEE International Conference on Computer
  Vision (ICCV)}, 3018--3027.

\bibitem[{He et~al.(2016)He, Zhang, Ren, and Sun}]{ResNet}
He, K.; Zhang, X.; Ren, S.; and Sun, J. 2016.
\newblock Deep residual learning for image recognition.
\newblock In \emph{Proceedings of the IEEE conference on computer vision and
  pattern recognition}, 770--778.

\bibitem[{Hu et~al.(2020)Hu, Moreno, Xiao, Shen, Obozinski, Lawrence, and
  Damianou}]{SIB}
Hu, S.~X.; Moreno, P.~G.; Xiao, Y.; Shen, X.; Obozinski, G.; Lawrence, N.; and
  Damianou, A. 2020.
\newblock Empirical Bayes Transductive Meta-Learning with Synthetic Gradients.
\newblock In \emph{International Conference on Learning Representations}.

\bibitem[{Islam et~al.(2021)Islam, Chen, Panda, Karlinsky, Radke, and
  Feris}]{Contrastive}
Islam, A.; Chen, C.-F.~R.; Panda, R.; Karlinsky, L.; Radke, R.; and Feris, R.
  2021.
\newblock A Broad Study on the Transferability of Visual Representations With
  Contrastive Learning.
\newblock In \emph{Proceedings of the IEEE/CVF International Conference on
  Computer Vision (ICCV)}, 8845--8855.

\bibitem[{Jamal and Qi(2019)}]{TAML}
Jamal, M.~A.; and Qi, G.-J. 2019.
\newblock Task Agnostic Meta-Learning for Few-Shot Learning.
\newblock In \emph{Proceedings of the IEEE/CVF Conference on Computer Vision
  and Pattern Recognition (CVPR)}, 11719--11727.

\bibitem[{Kang et~al.(2021)Kang, Kwon, Min, and Cho}]{RENet}
Kang, D.; Kwon, H.; Min, J.; and Cho, M. 2021.
\newblock Relational Embedding for Few-Shot Classification.
\newblock In \emph{Proceedings of the IEEE/CVF International Conference on
  Computer Vision (ICCV)}, 8822--8833.

\bibitem[{Koch et~al.(2015)Koch, Zemel, Salakhutdinov et~al.}]{FSL3}
Koch, G.; Zemel, R.; Salakhutdinov, R.; et~al. 2015.
\newblock Siamese neural networks for one-shot image recognition.
\newblock In \emph{ICML deep learning workshop}, volume~2.

\bibitem[{Krizhevsky, Sutskever, and Hinton(2012)}]{ObjectRecognition}
Krizhevsky, A.; Sutskever, I.; and Hinton, G.~E. 2012.
\newblock ImageNet Classification with Deep Convolutional Neural Networks.
\newblock In \emph{Advances in Neural Information Processing Systems},
  volume~25. Curran Associates, Inc.

\bibitem[{Lake, Salakhutdinov, and Tenenbaum(2015)}]{FSL2}
Lake, B.~M.; Salakhutdinov, R.; and Tenenbaum, J.~B. 2015.
\newblock Human-level concept learning through probabilistic program induction.
\newblock \emph{Science}, 350(6266): 1332--1338.

\bibitem[{LeCun, Bengio, and Hinton(2015)}]{DeepLearning}
LeCun, Y.; Bengio, Y.; and Hinton, G. 2015.
\newblock Deep learning.
\newblock \emph{nature}, 521(7553): 436--444.

\bibitem[{Lee, Hwang, and Shin(2020)}]{SLA}
Lee, H.; Hwang, S.~J.; and Shin, J. 2020.
\newblock Self-supervised Label Augmentation via Input Transformations.
\newblock In \emph{Proceedings of the 37th International Conference on Machine
  Learning}, volume 119 of \emph{Proceedings of Machine Learning Research},
  5714--5724. PMLR.

\bibitem[{Lee et~al.(2019)Lee, Maji, Ravichandran, and Soatto}]{MetaOptNet}
Lee, K.; Maji, S.; Ravichandran, A.; and Soatto, S. 2019.
\newblock Meta-Learning With Differentiable Convex Optimization.
\newblock In \emph{Proceedings of the IEEE/CVF Conference on Computer Vision
  and Pattern Recognition (CVPR)}, 10657--10665.

\bibitem[{Li, Fergus, and Perona(2006)}]{FSL1}
Li, F.-F.; Fergus, R.; and Perona, P. 2006.
\newblock One-shot learning of object categories.
\newblock \emph{IEEE Transactions on Pattern Analysis and Machine
  Intelligence}, 28(4): 594--611.

\bibitem[{Li et~al.(2019)Li, Eigen, Dodge, Zeiler, and Wang}]{CTM}
Li, H.; Eigen, D.; Dodge, S.; Zeiler, M.; and Wang, X. 2019.
\newblock Finding Task-Relevant Features for Few-Shot Learning by Category
  Traversal.
\newblock In \emph{Proceedings of the IEEE/CVF Conference on Computer Vision
  and Pattern Recognition (CVPR)}, 1--10.

\bibitem[{Li et~al.(2021)Li, He, Zhang, Liu, Zhang, and Wu}]{Person}
Li, Y.; He, J.; Zhang, T.; Liu, X.; Zhang, Y.; and Wu, F. 2021.
\newblock Diverse Part Discovery: Occluded Person Re-Identification With
  Part-Aware Transformer.
\newblock In \emph{Proceedings of the IEEE/CVF Conference on Computer Vision
  and Pattern Recognition (CVPR)}, 2898--2907.

\bibitem[{Li et~al.(2017)Li, Zhou, Chen, and Li}]{Meta-SGD}
Li, Z.; Zhou, F.; Chen, F.; and Li, H. 2017.
\newblock Meta-SGD: Learning to Learn Quickly for Few Shot Learning.
\newblock \emph{CoRR}, abs/1707.09835.

\bibitem[{Liu et~al.(2020)Liu, Cao, Lin, Li, Zhang, Long, and Hu}]{Neg-Cosine}
Liu, B.; Cao, Y.; Lin, Y.; Li, Q.; Zhang, Z.; Long, M.; and Hu, H. 2020.
\newblock Negative Margin Matters: Understanding Margin in Few-Shot
  Classification.
\newblock In \emph{Computer Vision -- ECCV 2020}, 438--455. Cham: Springer
  International Publishing.

\bibitem[{Liu et~al.(2021)Liu, Fu, Xu, Yang, Li, Wang, and Zhang}]{InfoPatch}
Liu, C.; Fu, Y.; Xu, C.; Yang, S.; Li, J.; Wang, C.; and Zhang, L. 2021.
\newblock Learning a Few-shot Embedding Model with Contrastive Learning.
\newblock In \emph{Proceedings of the AAAI Conference on Artificial
  Intelligence}, volume~35, 8635--8643.

\bibitem[{Liu, Schiele, and Sun(2020)}]{E3BM}
Liu, Y.; Schiele, B.; and Sun, Q. 2020.
\newblock An Ensemble of Epoch-Wise Empirical Bayes for Few-Shot Learning.
\newblock In \emph{Computer Vision -- ECCV 2020}, 404--421. Springer
  International Publishing.

\bibitem[{Meng et~al.(2021)Meng, Zhao, Huang, and Zhou}]{FaceRecognition}
Meng, Q.; Zhao, S.; Huang, Z.; and Zhou, F. 2021.
\newblock MagFace: A Universal Representation for Face Recognition and Quality
  Assessment.
\newblock In \emph{Proceedings of the IEEE/CVF Conference on Computer Vision
  and Pattern Recognition (CVPR)}, 14225--14234.

\bibitem[{Ouali, Hudelot, and Tami(2021)}]{SCL}
Ouali, Y.; Hudelot, C.; and Tami, M. 2021.
\newblock Spatial Contrastive Learning for Few-Shot Classification.
\newblock In \emph{Joint European Conference on Machine Learning and Knowledge
  Discovery in Databases}, 671--686. Springer International Publishing.

\bibitem[{Park et~al.(2020)Park, Han, Baek, Kim, Song, Lee, Han, and
  Hwang}]{Meta-Variance}
Park, S.-J.; Han, S.; Baek, J.-W.; Kim, I.; Song, J.; Lee, H.~B.; Han, J.-J.;
  and Hwang, S.~J. 2020.
\newblock Meta Variance Transfer: Learning to Augment from the Others.
\newblock In \emph{Proceedings of the 37th International Conference on Machine
  Learning}, volume 119 of \emph{Proceedings of Machine Learning Research},
  7510--7520. PMLR.

\bibitem[{Rajasegaran et~al.(2021)Rajasegaran, Khan, Hayat, Khan, and
  Shah}]{SKD}
Rajasegaran, J.; Khan, S.; Hayat, M.; Khan, F.~S.; and Shah, M. 2021.
\newblock Self-supervised Knowledge Distillation for Few-shot Learning.
\newblock In \emph{British Machine Vision Conference}.

\bibitem[{Ravi and Larochelle(2017)}]{Meta-LSTM}
Ravi, S.; and Larochelle, H. 2017.
\newblock Optimization as a Model for Few-Shot Learning.
\newblock In \emph{International Conference on Learning Representations}.

\bibitem[{Ren et~al.(2018)Ren, Ravi, Triantafillou, Snell, Swersky, Tenenbaum,
  Larochelle, and Zemel}]{tieredImageNet}
Ren, M.; Ravi, S.; Triantafillou, E.; Snell, J.; Swersky, K.; Tenenbaum, J.~B.;
  Larochelle, H.; and Zemel, R.~S. 2018.
\newblock Meta-Learning for Semi-Supervised Few-Shot Classification.
\newblock In \emph{International Conference on Learning Representations}.

\bibitem[{Rizve et~al.(2021)Rizve, Khan, Khan, and Shah}]{Inv-Equ}
Rizve, M.~N.; Khan, S.; Khan, F.~S.; and Shah, M. 2021.
\newblock Exploring Complementary Strengths of Invariant and Equivariant
  Representations for Few-Shot Learning.
\newblock In \emph{Proceedings of the IEEE/CVF Conference on Computer Vision
  and Pattern Recognition (CVPR)}, 10836--10846.

\bibitem[{Rusu et~al.(2019)Rusu, Rao, Sygnowski, Vinyals, Pascanu, Osindero,
  and Hadsell}]{LEO}
Rusu, A.~A.; Rao, D.; Sygnowski, J.; Vinyals, O.; Pascanu, R.; Osindero, S.;
  and Hadsell, R. 2019.
\newblock Meta-Learning with Latent Embedding Optimization.
\newblock In \emph{International Conference on Learning Representations}.

\bibitem[{Satorras and Estrach(2018)}]{GNN}
Satorras, V.~G.; and Estrach, J.~B. 2018.
\newblock Few-Shot Learning with Graph Neural Networks.
\newblock In \emph{International Conference on Learning Representations}.

\bibitem[{Schwartz et~al.(2018)Schwartz, Karlinsky, Shtok, Harary, Marder,
  Kumar, Feris, Giryes, and Bronstein}]{Delta-encoder}
Schwartz, E.; Karlinsky, L.; Shtok, J.; Harary, S.; Marder, M.; Kumar, A.;
  Feris, R.; Giryes, R.; and Bronstein, A. 2018.
\newblock Delta-encoder: an effective sample synthesis method for few-shot
  object recognition.
\newblock In \emph{Advances in Neural Information Processing Systems},
  volume~31. Curran Associates, Inc.

\bibitem[{Simon et~al.(2020)Simon, Koniusz, Nock, and Harandi}]{DSN}
Simon, C.; Koniusz, P.; Nock, R.; and Harandi, M. 2020.
\newblock Adaptive Subspaces for Few-Shot Learning.
\newblock In \emph{Proceedings of the IEEE/CVF Conference on Computer Vision
  and Pattern Recognition (CVPR)}, 4136--4145.

\bibitem[{Snell, Swersky, and Zemel(2017)}]{ProtoNet}
Snell, J.; Swersky, K.; and Zemel, R. 2017.
\newblock Prototypical Networks for Few-shot Learning.
\newblock In \emph{Advances in Neural Information Processing Systems},
  volume~30. Curran Associates, Inc.

\bibitem[{Su, Maji, and Hariharan(2020)}]{Proto-Jigsaw}
Su, J.-C.; Maji, S.; and Hariharan, B. 2020.
\newblock When Does Self-supervision Improve Few-Shot Learning?
\newblock In \emph{Computer Vision -- ECCV 2020}, 645--666. Springer
  International Publishing.

\bibitem[{Sung et~al.(2018)Sung, Yang, Zhang, Xiang, Torr, and
  Hospedales}]{RelationNet}
Sung, F.; Yang, Y.; Zhang, L.; Xiang, T.; Torr, P.~H.; and Hospedales, T.~M.
  2018.
\newblock Learning to Compare: Relation Network for Few-Shot Learning.
\newblock In \emph{Proceedings of the IEEE Conference on Computer Vision and
  Pattern Recognition (CVPR)}, 1199--1208.

\bibitem[{Tang et~al.(2021)Tang, Chen, Bai, Liu, Ge, and Ouyang}]{CRF-GNN}
Tang, S.; Chen, D.; Bai, L.; Liu, K.; Ge, Y.; and Ouyang, W. 2021.
\newblock Mutual CRF-GNN for Few-Shot Learning.
\newblock In \emph{Proceedings of the IEEE/CVF Conference on Computer Vision
  and Pattern Recognition (CVPR)}, 2329--2339.

\bibitem[{Tian et~al.(2020)Tian, Wang, Krishnan, Tenenbaum, and Isola}]{RFS}
Tian, Y.; Wang, Y.; Krishnan, D.; Tenenbaum, J.~B.; and Isola, P. 2020.
\newblock Rethinking Few-Shot Image Classification: A Good Embedding is All You
  Need?
\newblock In \emph{Computer Vision -- ECCV 2020}, 266--282. Springer
  International Publishing.

\bibitem[{Vinyals et~al.(2016)Vinyals, Blundell, Lillicrap, kavukcuoglu, and
  Wierstra}]{MatchingNet}
Vinyals, O.; Blundell, C.; Lillicrap, T.; kavukcuoglu, k.; and Wierstra, D.
  2016.
\newblock Matching Networks for One Shot Learning.
\newblock In \emph{Advances in Neural Information Processing Systems},
  volume~29. Curran Associates, Inc.

\bibitem[{Wah et~al.(2011)Wah, Branson, Welinder, Perona, and Belongie}]{CUB}
Wah, C.; Branson, S.; Welinder, P.; Perona, P.; and Belongie, S. 2011.
\newblock The Caltech-UCSD Birds-200-2011 Dataset.
\newblock Technical Report CNS-TR-2011-001, California Institute of Technology.

\bibitem[{Wang et~al.(2018)Wang, Girshick, Hebert, and Hariharan}]{Imaginary}
Wang, Y.-X.; Girshick, R.; Hebert, M.; and Hariharan, B. 2018.
\newblock Low-Shot Learning From Imaginary Data.
\newblock In \emph{Proceedings of the IEEE Conference on Computer Vision and
  Pattern Recognition (CVPR)}, 7278--7286.

\bibitem[{Wertheimer, Tang, and Hariharan(2021)}]{FRN}
Wertheimer, D.; Tang, L.; and Hariharan, B. 2021.
\newblock Few-Shot Classification With Feature Map Reconstruction Networks.
\newblock In \emph{Proceedings of the IEEE/CVF Conference on Computer Vision
  and Pattern Recognition (CVPR)}, 8012--8021.

\bibitem[{Wu et~al.(2021)Wu, Zhang, Zhang, and Wu}]{TPMN}
Wu, J.; Zhang, T.; Zhang, Y.; and Wu, F. 2021.
\newblock Task-Aware Part Mining Network for Few-Shot Learning.
\newblock In \emph{Proceedings of the IEEE/CVF International Conference on
  Computer Vision (ICCV)}, 8433--8442.

\bibitem[{Yang, Liu, and Xu(2021)}]{Free-Lunch}
Yang, S.; Liu, L.; and Xu, M. 2021.
\newblock Free Lunch for Few-shot Learning: Distribution Calibration.
\newblock In \emph{International Conference on Learning Representations}.

\bibitem[{Ye et~al.(2020)Ye, Hu, Zhan, and Sha}]{FEAT}
Ye, H.-J.; Hu, H.; Zhan, D.-C.; and Sha, F. 2020.
\newblock Few-Shot Learning via Embedding Adaptation With Set-to-Set Functions.
\newblock In \emph{Proceedings of the IEEE/CVF Conference on Computer Vision
  and Pattern Recognition (CVPR)}, 8808--8817.

\bibitem[{Zhang et~al.(2020)Zhang, Cai, Lin, and Shen}]{DeepEMD}
Zhang, C.; Cai, Y.; Lin, G.; and Shen, C. 2020.
\newblock DeepEMD: Few-Shot Image Classification With Differentiable Earth
  Mover's Distance and Structured Classifiers.
\newblock In \emph{Proceedings of the IEEE/CVF Conference on Computer Vision
  and Pattern Recognition (CVPR)}, 12203--12213.

\bibitem[{Zhang et~al.(2021)Zhang, Zhang, Lu, Xiang, Ding, and Huang}]{IEPT}
Zhang, M.; Zhang, J.; Lu, Z.; Xiang, T.; Ding, M.; and Huang, S. 2021.
\newblock IEPT: Instance-Level and Episode-Level Pretext Tasks for Few-Shot
  Learning.
\newblock In \emph{International Conference on Learning Representations}.

\bibitem[{Zhang et~al.(2018)Zhang, Che, Ghahramani, Bengio, and Song}]{MetaGAN}
Zhang, R.; Che, T.; Ghahramani, Z.; Bengio, Y.; and Song, Y. 2018.
\newblock MetaGAN: An Adversarial Approach to Few-Shot Learning.
\newblock In \emph{Advances in Neural Information Processing Systems},
  volume~31. Curran Associates, Inc.

\end{thebibliography}

\end{document}